\documentclass[letterpaper, 10 pt, conference]{ieeeconf}  % Comment this line out if you need a4paper

\IEEEoverridecommandlockouts                              % This command is only needed if 
\usepackage{amsmath}
\usepackage{graphics} % for pdf, bitmapped graphics files
\usepackage{epsfig} % for postscript graphics files
\usepackage[noend]{algpseudocode}
\usepackage{algorithm}
\usepackage{rotating}
\usepackage{makecell}
\usepackage{multirow}
\usepackage{cite}
\usepackage[ruled,vlined,algo2e]{algorithm2e}

\usepackage[breaklinks=true, colorlinks, bookmarks=false]{hyperref}
\usepackage{url}
\usepackage{amssymb}% http://ctan.org/pkg/amssymb
\usepackage{pifont}% http://ctan.org/pkg/pifont
\newcommand{\cmark}{\ding{51}}%
\newcommand{\xmark}{\ding{55}}%

\title{\LARGE \bf
RangeSeg: Efficient Lidar Semantic Segmentation on Range view
}

\author{Ji-Chao Jiao*, Ben Ding, Ning Li, Min Pang
}

\begin{document}

\maketitle
\thispagestyle{empty}
\pagestyle{empty}

%%%%%%%%%%%%%%%%%%%%%%%%%%%%%%%%%%%%%%%%%%%%%%%%%%%%%%%%%%%%%%%%%%%%%%%%%%%%%%%%
\begin{abstract}
LiDAR-based semantic segmentation is critical in the fields of robotics and autonomous driving as it provides a comprehensive understanding of the scene. This paper proposes a lightweight and efficient projection-based semantic segmentation network called LENet with an encoder-decoder structure for LiDAR-based semantic segmentation. The encoder is composed of a novel multi-scale convolutional attention (MSCA) module with varying receptive field sizes to capture features. The decoder employs an Interpolation And Convolution (IAC) mechanism utilizing bilinear interpolation for upsampling multi-resolution feature maps and integrating previous and current dimensional features through a single convolution layer. This approach significantly reduces the network's complexity while also improving its accuracy. Additionally, we introduce multiple auxiliary segmentation heads to further refine the network's accuracy. Extensive evaluations on publicly available datasets, including SemanticKITTI, SemanticPOSS, and nuScenes, show that our proposed method is lighter, more efficient, and robust compared to state-of-the-art semantic segmentation methods. Full implementation is available at \url{https://github.com/fengluodb/LENet}.
\end{abstract}

\section{Introduction}
\begin{figure}[ht]
	\centering
	\begin{minipage}{1.0\linewidth}
		\centering
            \setlength{\abovecaptionskip}{-0.2cm}
		\includegraphics[width=1.0\linewidth]{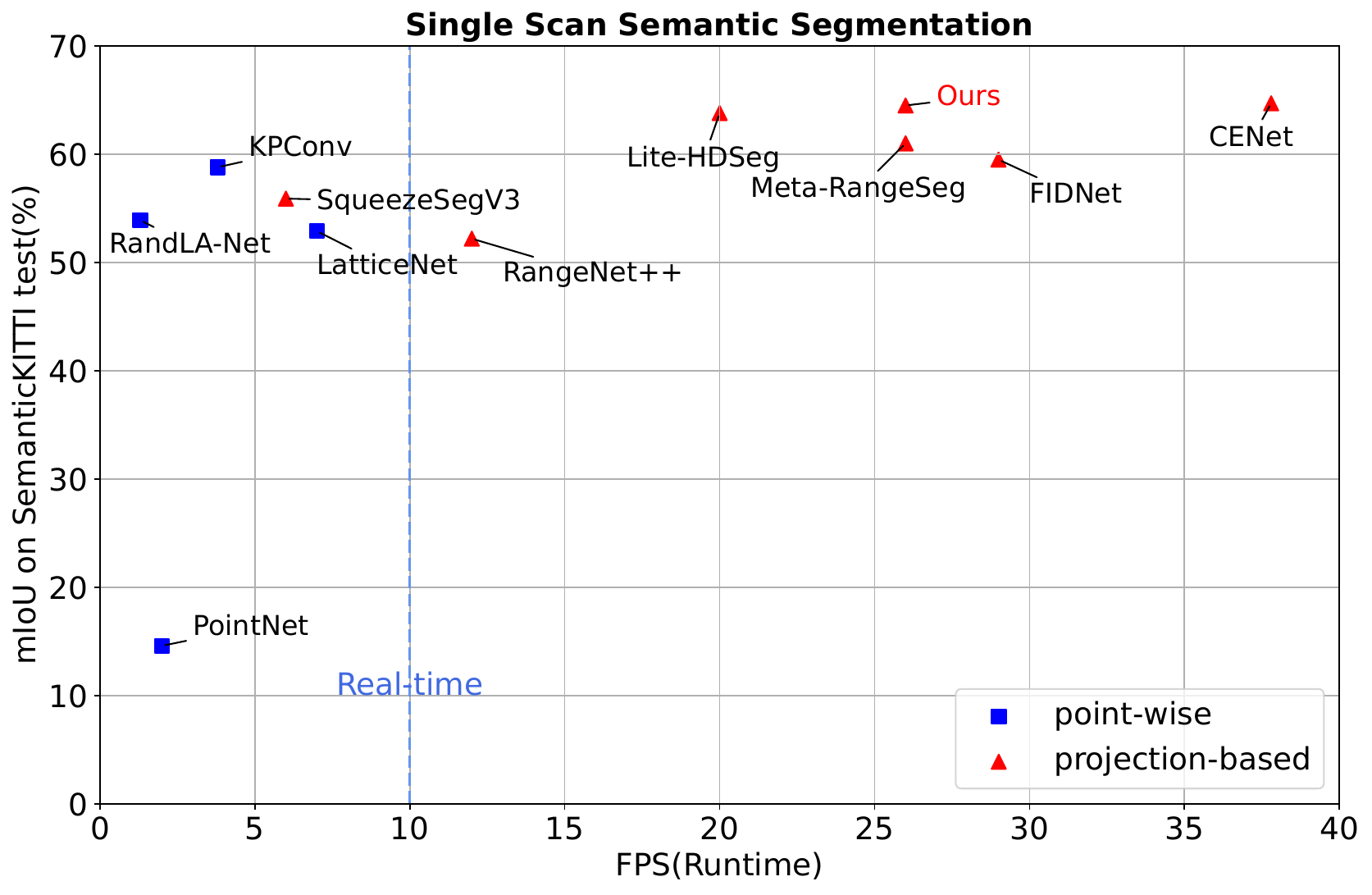}
		\label{fig:fig1a}%文中引用该图片代号
            \vspace{-5mm}
	\end{minipage}
	%\qquad
	
	\begin{minipage}{1.0\linewidth}
		\centering
            \setlength{\abovecaptionskip}{-0.2cm}
		\includegraphics[width=1.0\linewidth]{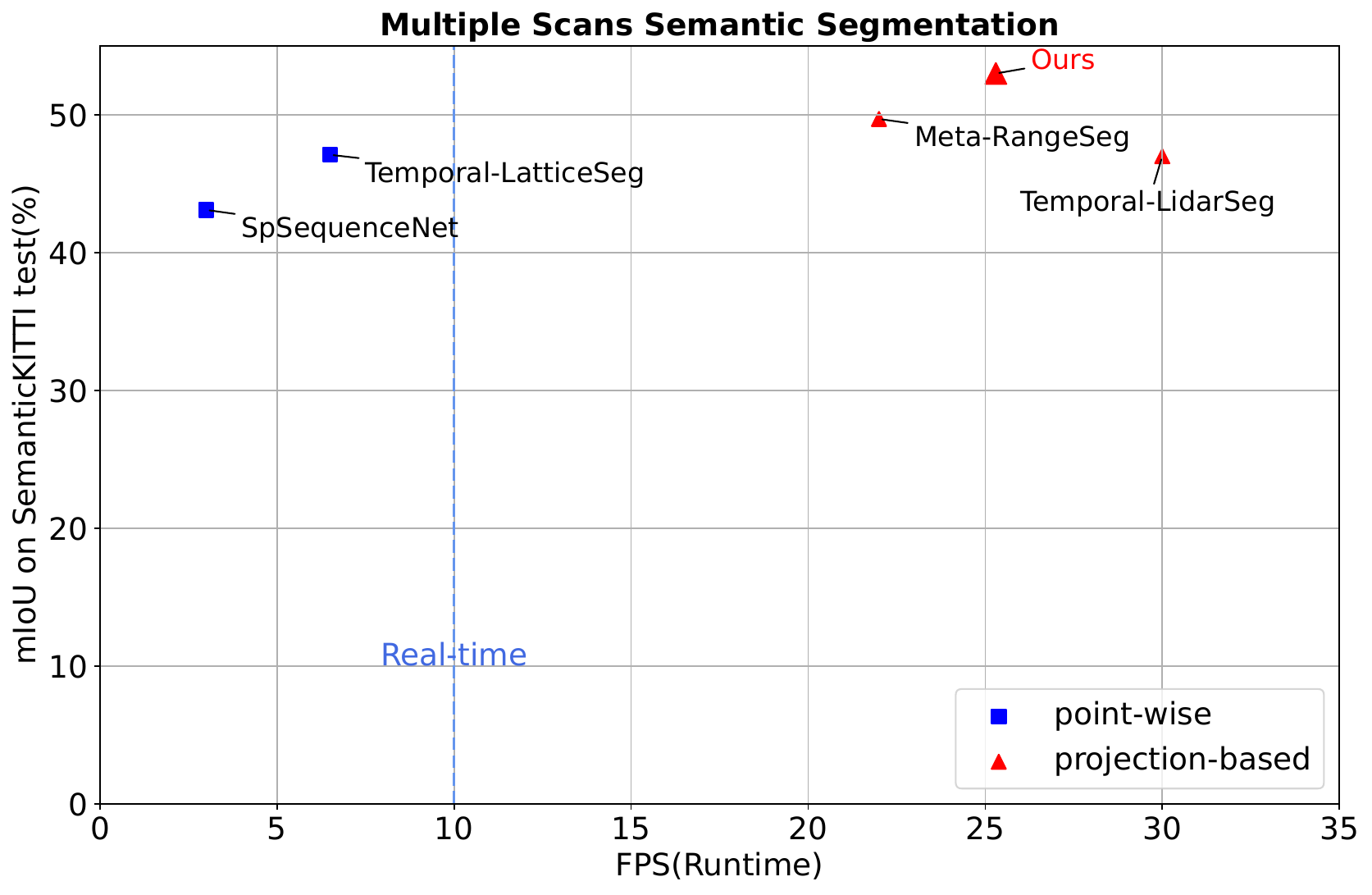}
		\caption{Accuracy(mIoU) versus inference time. Our presented LENet obtains the promising results on both single scan and multiple scans semantic segmentation in SemanticKITTI test dataset~\cite{behley2019semantickitti}.}
		\label{fig:fig1b}%文中引用该图片代号
            \vspace{-6mm}
	\end{minipage}
\end{figure}

% \begin{figure}
% \centering
% \setlength{\abovecaptionskip}{-0.2cm}
% \includegraphics[width=1.0\linewidth,height=50mm]{fig/singe.pdf}
%     \vspace{-6mm}
% \end{figure}

% \begin{figure}
% \centering
% \setlength{\abovecaptionskip}{-0.2cm}
% \includegraphics[width=1.0\linewidth,height=50mm]{fig/multi.pdf}
%     \caption{Accuracy(mIoU) versus inference time. Our presented LENet obtains the promising results on both single scan and multiple scans semantic segmentation in SemanticKITTI test dataset~\cite{behley2019semantickitti}.}
%     \label{fig:fig1}
%     \vspace{-6mm}
% \end{figure}
Environment perception is crucial for autonomous driving vehicles to comprehend the surrounding scene. LiDAR and RGB cameras are commonly employed in the perception system of autonomous driving. While cameras have their advantages, such as high resolution and low cost, LiDAR sensors are more robust and not affected by lighting and weather conditions. Additionally, LiDAR point cloud data provides a geometry-accurate representation of objects, making it an ideal choice for 3D point cloud analysis. Point cloud semantic segmentation, which assigns labels to individual points in a point cloud, provides a rich understanding of the scene. Consequently, point cloud semantic segmentation has become an increasingly popular research topic in both academic and industrial communities.

Due to the disorder and irregularity of 3D point clouds, it is not feasible to apply standard convolutional neural networks directly. Therefore, research efforts have been focused on exploring point cloud semantic segmentation. Point-based methods, including PointNet~\cite{qi2017pointnet}, PointNet++~\cite{qi2017pointnet++}, PointCNN~\cite{li2018pointcnn}, and RandLA-Net~\cite{hu2020randla}, directly extract features from raw point clouds, which reduces the impact of computational complexity and noise errors from the pre-processing stage. However, point-based methods suffer from high computational complexity and limited processing speed. Voxel-based methods convert the irregular point clouds into regular grid representations, which allows them to use 3D convolutional neural networks. However, voxel-based methods still face problems similar to point-based methods. Projection-based methods overcome these issues by transforming point clouds into 2D range images via spherical projection. In addition to delivering superior inference speed and accuracy compared to point-based and voxel-based methods, projection-based methods have recently become increasingly popular due to the success of convolutional networks in image semantic segmentation.

In this paper, we propose a lightweight and efficient LiDAR semantic segmentation network that utilizes projection-based methods. Through extensive experimentation on popular public benchmarks, such as SemanticKITTI~\cite{behley2019semantickitti}, we have demonstrated that our network achieves state-of-the-art accuracy with a speed of 26 frames per second (as shown in Figure~\ref{fig:fig1a}). In summary, the main contributions of this paper are as follows:

\begin{itemize}
\setlength{\itemsep}{0pt}
\setlength{\parsep}{0pt}
\setlength{\parskip}{0pt}
\item We have developed a novel multi-scale convolutional attention module (MSCA) to replace the ResNet block within our encoder. By using varying kernel sizes (3,5,7), MSCA is able to capture critical information about objects of differing sizes within the LiDAR scan. This component plays a vital role in optimizing semantic segmentation.
\item We have introduced a novel IAC module that combines bilinear interpolation for upsampling multi-resolution feature maps with a single convolution layer for integrating previous and current features. Our approach results in a lightweight and powerful mechanism that significantly reduces the complexity of the decoder while also improving accuracy.
\item We increased the accuracy of the network without adding additional inference parameters by incorporating multiple auxiliary segmentation heads. These segmentation heads refine feature maps at different resolutions, ultimately improving the learning capabilities of the network.
\item We performed comprehensive experiments on publicly available datasets, including SemanticKITTI, SemanticPOSS, and nuScenes. The results indicate that our proposed method attains state-of-the-art performance.
\end{itemize}

\begin{figure*}
\centering
\setlength{\abovecaptionskip}{-0.1cm}
\includegraphics[width=1.0\linewidth,height=45mm]{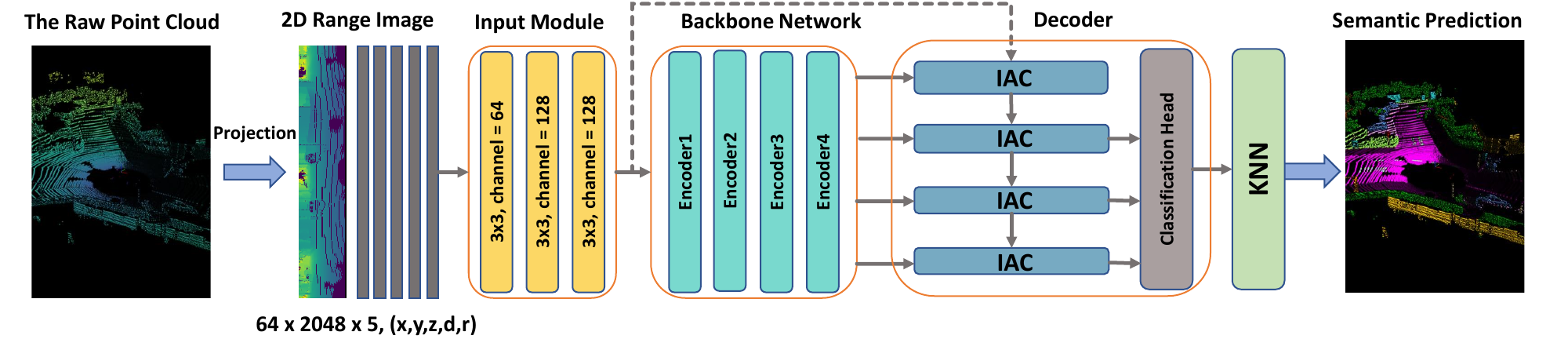}
    \caption{Illustration of our proposed LENet framework. The backbone network is consisted of MSCA and has the pyramid structure similar to ResNet34. The ICA module upsamples the low-dimensional feature maps to original size and aggregate it with the output of the previous IAC module. Then, the last classification head receives the feature maps from the last three ICA module and outputs the label of each point. Finally, we use K-nearest Neighbors (KNN)\cite{milioto2019rangenet++} to achieve post-processing.} 
    \label{fig:second}
    \vspace{-4mm}
    % \vspace{-2.8cm}
\end{figure*}

\section{RELATED WORK}

In recent years, the availability of large-scale datasets such as SemanticKITTI~\cite{behley2019semantickitti}, NuScenes~\cite{nuscenes2019}, and SemanticPOSS~\cite{pan2020semanticposs} for the task of point cloud segmentation of autonomous driving scenes, combined with the rapid development of deep learning, has led to the proposal of a wide range of 3D LiDAR point cloud semantic segmentation methods. These methods can generally be categorized into four groups based on their input data representation, including point, voxel, range map, and hybrid representations.

\textbf{Point-based methods} directly process the raw 3D point cloud without applying any additional transformation or pre-processing, which are able to preserve the 3D spatial structure information. The pioneering methods of this group are PointNet\cite{qi2017pointnet} and PointNet++\cite{qi2017pointnet++}, which use shared MLPs\cite{mcclelland1987parallel} to learn the properties of each point. In subsequent series of works, KPConv\cite{thomas2019kpconv} develops deformable convolutions that can use arbitrary number of kernel points to learn local geometry.  Howerver, these approaches have the disadvantages of high computational complexity and large memory consumption, which hinders them from the large-scale point cloud. RandLA-Net\cite{hu2020randla} adopts a random sampling strategy and uses local feature aggregation to reduce the information loss caused by random operations, which considerably improve the efficiency of point cloud processing and decrease the use of memory consumption.

\textbf{Voxel-based Methods} Voxel-based approach to convert point clouds into voxels for processing, which can effectively solve the irregularity problem. The early voxel-based methods firstly transform a point cloud into 3D voxel representations, then use the standard 3D CNN to predict semantic labels. However, the regular 3D convolution requires the huge memory and heavy computational power. Minkowski\cite{choy20194d} chose to use sparse convolution instead of standard 3D convolution and other standard neural network to reduce the computational cost. Cylinder3D\cite{zhu2020cylindrical} adopts 3D space partition and designs an asymmetrical residual block to reduce computation. AF2S3Net\cite{cheng20212} achieves state-of-the-art of voxel-methods, which proposes two novel attention blocks name Attentive Feature Fustion Module(AF2M) and Adaptive Feature Fustion(ASFM) to effectively learn local features and global contexts.

\textbf{Projection-based Methods} project 3D point clouds into 2D image space, which can take advantage of a large amount of advanced layers for image feature extraction. SqueezeSeg\cite{wu2018squeezeseg} proposes spherical projection which maps the scatter 3D laser points into 2D Range-Image, then uses the lightweight model SqueezeNet and CRF for segmentation. Subsequently, SqueezeSegV2\cite{wu2019squeezesegv2} proposes context aggregation module (CAM) to aggregate contextual information from a larger perceptual field. RangeNet++\cite{milioto2019rangenet++} integrates Darknet into SqueezeSeg and proposes an efficient KNN post-processing method to predict labels for point. SqueezeSegV3\cite{xu2020squeezesegv3} proposes Spatially-Adaptive Convolution (SAC) with different filters depending on the location of the input image. SalsaNext\cite{cortinhal2020salsanext} inherits the encoder-decoder architecture from SalsaNet\cite{aksoy2019salsanet} and presents an uncertainty-aware mechanism for point feature leaning. Lite-HDSeg\cite{razani2021lite} achieves state-of-the-art performance by introducing three different modules, Inception-like Context Module, Multi-class Spatial Propagation Network, and a boundary loss. To get better performance and inspired by the success of SegNext on image segmentation, we propose a novel MSCA module and IAC module to compose the new pipeline on the basis of FIDNet\cite{zhao2021fidnet}.

\textbf{Hybrid methods} try to integrate the advantages of the point-wise methods, the projection methods and voxel methods. SPVNAS\cite{tang2020searching} proposed a sparse point-voxel convolution, which consists of the point-based branch and sparse voxel-based branch. The point-based branch focuses on the local features of the point cloud, the sparse voxel-based branch focuses on global contexts. RPVNet\cite{xu2021rpvnet} propose a network that fuses three types of data, which includes three branches: point-base branch, projection-based branch and voxel-based branch. Point-based branch acts as an intermediate node to fuse the output of different branches. 

\section{METHOD} \label{method}

\subsection{Range Image Representation.} 
Using the spherical projection approach, we can transform the unstructured point cloud into an ordered range representation that is a 2D image. The advantages of the range representation are that it can use the effective 2D convolutional operation for fast training and inference, and it can facilitate the mature deep learning technologies that have been well studied in image-based tasks. 

In the range image representation, each LiDAR point ${p}=(x, y, z)$ with Cartesian coordinates, a spherical mapping $\mathbb{R}^3 \rightarrow \mathbb{R}^2$ is used to transform it to image coordinates, as below:
\begin{align}
	\left( \begin{array}{c} u \vspace{0.0em} \\ v \end{array}\right) & = \left(\begin{array}{cc} \frac{1}{2}\left[1-\arctan(y, x) \, \pi^{-1}\right]~\,~w   \vspace{0.5em} \\
			\left[1 - \left(\arcsin(z\, r^{-1}) + \mathrm{f}_{\mathrm{up}}\right) \mathrm{f}^{-1}\right] \, h\end{array} \right), \label{eq:projection}
\end{align}
where $(u,v)$ are image coordinates, $(h, w)$ are the height and width of the desired range image representation, $\mathrm{f}~{=}~\mathrm{f}_{\mathrm{up}}~{+}~\mathrm{f}_{\mathrm{down}}$ is the vertical field-of-view of the sensor, and $r~{=}~\sqrt{x^2+y^2+z^2}$ is the range of each point.

\subsection{Convolution Attention Encoder}
Multi-scale features are crucial in semantic segmentation as they enable the processing of objects of varying sizes in a single image. To extract these features, a conventional method involves using convolutions with different receptive fields and fusing their respective fields as shown in~\cite{cortinhal2020salsanext}. Inspired by SegNext~\cite{guo2022segnext}, we propose a novel multi-scale convolution attention (MSCA) module customized for our problem domain. In contrast to the original version in SegNext, we have made significant modifications and adaptations. These include resizing the MSCN convolutional kernel to suit lidar data, replacing the GELU activation function with SiLU to enhance efficiency while maintaining accuracy, and optimizing the BasicBlock structure to fully utilize MSCN's capabilities.

\begin{figure}
 \centering
 \setlength{\abovecaptionskip}{0.2cm}
\includegraphics[width=1.0\linewidth,height=50mm]{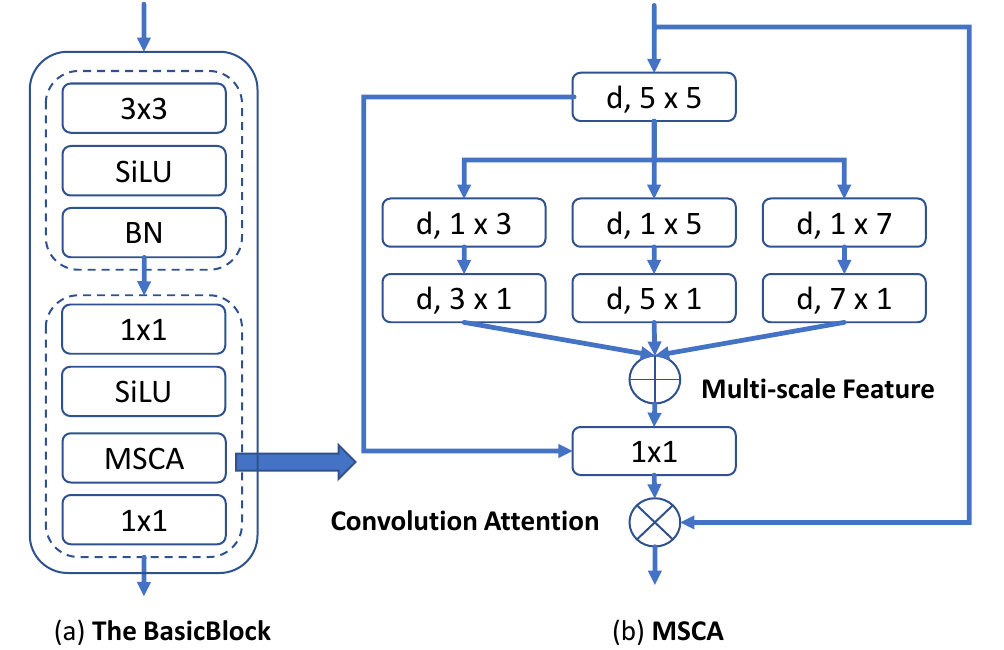}
    \caption{Illustration of the  BasicBlock that is used to build the encoder and the proposed MSCA. Here, $d$ means a depth-wise convolution, $k_1 \times k_2$ means the kernel size of the convolution layer. We extract multi-scale features using the convolutions and then utilize them as attention weights to reweight the input of MSCA. }
    \label{fig:third}
    \vspace{-6mm}
\end{figure}

As shown in Fig. \ref{fig:third} (b), the MSCA module comprises three parts: a depth-wise convolution to aggregate local information, multi-branch depth-wise strip convolutions to capture multi-scale context, and a $1 \times 1$ convolution to model the relationship between different channels. Finally, the output of the $1\times1$ convolution is utilized directly as attention weights to re-weight the input of MSCA. Moreover, we follow ~\cite{zhao2021fidnet} in adopting the pyramid structure for our encoder. The building block in the encoder, as depicted in Fig.\ref{fig:third} (a), consists of a $3\times3$ convolution layer and MSCA.

\subsection{IAC Decoder}
To design a simple and effective decoder, we investigate several different decoder structure. In ~\cite{milioto2019rangenet++, xu2020squeezesegv3, cortinhal2020salsanext}, they use standard transposed convolutions or \textit{pixel-shuffle} to produce the upsampled feature maps, then using a set of convolution to decoder the feature maps, which is effective but with heavy computation. In FIDNet, it's decoder uses FID (fully interpolation decoding) to decode the semantics of different levels, then using a classification head to fuse these semantics. Although FID is completely parameter-free, it doesn't have the ability to learn from the features, which makes the model's performance excessively depend on the classification head. Besides, FIDNet's classification head fuses too much low-level information, which decreases the performance. 

In this work, we presented a lightweight decoder as depicted in Fig.~\ref{fig:second}. The IAC module contains two parts: a bilinear interpolation to upsample the feature maps which come from the encoder, and a $3\times3$ convolution to fuse the information from the encoder and the previous IAC. Finally, we use the point-wise convolution to fuse the features from the last three IAC modules. Comparing to other two decoders, our decoder has the fewest parameters and the best performance.

\subsection{Loss Function}
\begin{table*}[ht!] 
\begin{center}
	\caption{The performance comparison on SemanticKITTI single scan benchmark. }
	% \vspace{-3mm}
	\label{table1}
        \setlength{\tabcolsep}{1.pt}
	\resizebox{\textwidth}{!}{
 \begin{tabular}{c| c| c|c| c | c c c c c c c c c c c c c c c c c c c} 
 \hline
 Methods & Size & \rotatebox{90}{\textbf{mean-IoU}} & \rotatebox{90}{ {\textbf{FPS (Hz)}}} &\rotatebox{90}{ {\textbf{Params(M)}}} & \rotatebox{90}{car}& \rotatebox{90}{bicycle}& \rotatebox{90}{motorcycle}& \rotatebox{90}{truck}& \rotatebox{90}{other-vehicle}& \rotatebox{90}{person}& \rotatebox{90}{bicyclist}& \rotatebox{90}{motorcyclist}& \rotatebox{90}{road}& \rotatebox{90}{parking}& \rotatebox{90}{sidewalk}& \rotatebox{90}{other-ground}& \rotatebox{90}{building}& \rotatebox{90}{fence}& \rotatebox{90}{vegetation}& \rotatebox{90}{trunk}& \rotatebox{90}{terrain}& \rotatebox{90}{pole}& \rotatebox{90}{traffic-sign} \\ 
 \hline\hline
 PointNet~\cite{qi2017pointnet} & 50K pts &14.6 &  {2} &3 &  46.3 &1.3 &0.3 &0.1& 0.8& 0.2& 0.2& 0.0& 61.6& 15.8& 35.7& 1.4& 41.4& 12.9& 31.0& 4.6& 17.6& 2.4& 3.7\\ 
 PointNet++~\cite{qi2017pointnet++} & 50K pts&20.1&  {0.1} &6 &53.7& 1.9& 0.2& 0.9& 0.2& 0.9& 1.0& 0.0& 72.0& 18.7& 41.8& 5.6& 62.3& 16.9& 46.5& 13.8& 30.0& 6.0 &8.9 \\
 SPLATNet~\cite{su2018splatnet} & 50K pts &22.8&  {1} &0.8 &66.6 &0.0& 0.0& 0.0& 0.0& 0.0& 0.0& 0.0& 70.4& 0.8& 41.5& 0.0& 68.7& 27.8& 72.3& 35.9 &35.8 &13.8 &0.0\\
 TangentConv~\cite{tatarchenko2018tangent} & 50K pts & 35.9 &  {0.3}&0.4 &86.8& 1.3& 12.7& 11.6& 10.2& 17.1& 20.2& 0.5 &82.9& 15.2& 61.7& 9.0& 82.8& 44.2& 75.5& 42.5& 55.5& 30.2& 22.2\\
 LatticeNet~\cite{rosu2019latticenet}& 50K pts &52.9&  {7} & -&92.9 &16.6& 22.2& 26.6& 21.4& 35.6& 43.0& 46.0& 90.0& 59.4& 74.1& 22.0& 88.2& 58.8& 81.7& 63.6& 63.1& 51.9& 48.4\\
  RandLA-Net~\cite{hu2020randla} & 50K pts& 53.9&  {1.3} &0.95 & 94.2& 26.0& 25.8& 40.1 & 38.9& 49.2 &48.2& 7.2& 90.7& 60.3& 73.7& 20.4& 86.9& 56.3& 81.4& 61.3& 66.8& 49.2& 47.7\\
  KPConv~\cite{thomas2019kpconv} & 50K pts &58.8&  {3.8} &14.9 &\textbf{96.0}& 30.2& 42.5& 33.4& 44.3 &61.5& 61.6& 11.8& 88.8& 61.3& 72.7& \textbf{31.6}& 90.5 & 64.2& \textbf{84.8} & {69.2}& {69.1}& 56.4& 47.4\\
  BAAF-Net~\cite{qiu2021semantic} & 50K pts & 59.9 &  {4.8}&1.23 & 95.4 & 31.8 & 35.5 & \textbf{48.7} & \textbf{46.7} & 49.5 & 55.7 & 33.0 & 90.9 & 62.2 & 74.4 & 23.6 & 89.8 & 60.8 & 82.7 & 63.4 & 67.9 & 53.7 &52.0 \\ 

 \hline
  
 RangeNet53++~\cite{milioto2019rangenet++}& $64\times 2048$&52.2&  {12}&50 & 91.4 &25.7& 34.4& 25.7& 23.0& 38.3& 38.8& 4.8& 91.8& 65.0& 75.2& 27.8& 87.4& 58.6& 80.5& 55.1& 64.6& 47.9& 55.9\\

 MINet~\cite{li2021multi} & $64\times 2048$& 55.2 &  {24}&- & 90.1 & 41.8 & 34.0 & 29.9 & 23.6 & 51.4 & 52.4 & 25.0 & 90.5 & 59.0 & 72.6 & 25.8 & 85.6 & 52.3 & 81.1 & 58.1 & 66.1 & 49.0 & 59.9  \\

  3D-MiniNet~\cite{alonso20203d} & $64\times 2048$&55.8 &  {28}&4 &90.5& 42.3& 42.1& 28.5& 29.4& 47.8& 44.1& 14.5& 91.6& 64.2& 74.5& 25.4& 89.4& 60.8& 82.8& 60.8& 66.7& 48.0& 56.6\\
 
  SqueezeSegV3~\cite{xu2020squeezesegv3}& $64\times 2048$&55.9& {6}&29.9&92.5& 38.7& 36.5& 29.6& 33.0& 45.6& 46.2& 20.1& 91.7& 63.4& 74.8& 26.4& 89.0& 59.4 &82.0& 58.7& 65.4& 49.6& 58.9\\

  SalsaNext~\cite{cortinhal2020salsanext}& $64\times 2048$&59.5&  {24}&6.7 & 91.9& 48.3& 38.6& 38.9& 31.9& 60.2& 59.0& 19.4& 91.7& 63.7& 75.8& 29.1& 90.2& 64.2 & 81.8& 63.6& 66.5 &54.3& 62.1 \\
  
   FIDNet~\cite{zhao2021fidnet} & $64\times 2048$&59.5&  \textbf{29} &6.0 &93.9&{54.7}&48.9&27.6&23.9&62.3&59.8&23.7&90.6&59.1&75.8&26.7&88.9&60.5&84.5&64.4&69.0&53.3&62.8\\

   Meta-RangeSeg~\cite{wang2022meta} & $64\times 2048$&61.0&  {26}&6.8 &93.9& 50.1 &43.8&43.9 &43.2&{63.7}&53.1&18.7&90.6&64.3&74.6&29.2&91.1&64.7&82.6&65.5&65.5&56.3&64.2\\

   Lite-HDSeg~\cite{razani2021lite} & $64\times 2048$&{63.8}&  {20}&-&92.3&40.0&\textbf{55.4}&37.7&39.6&59.2&\textbf{71.6}&\textbf{54.1}&\textbf{93.0}&68.2&\textbf{78.3}&29.3&\textbf{91.5}&{65.0}&78.2&65.8&65.1&{59.5}&\textbf{67.7}\\

   CENet~\cite{cheng2022cenet} & $64\times 2048$&\textbf{64.7}&  \textbf{37.8}& 6.7 &91.9&\textbf{58.6}&50.3&40.6&42.3&\textbf{68.9}&65.9&43.5&90.3&60.9&75.1&31.5&91.0&\textbf{66.2}&84.5&\textbf{69.7}&\textbf{70.0}&\textbf{61.5}&67.6\\
 \hline
 
 LENet(Ours) & $64\times 2048$&{64.5}& 26&  {4.7} &93.9& {57.0} &51.3&44.3 &44.4&{66.6}&64.9&36.0&91.8&\textbf{68.3}&76.9&{30.5}&{91.2}&{66.0}&83.7&{68.3}&{67.8}&{58.6}&63.2\\
 \hline
\end{tabular}}
\end{center}
\vspace{-5mm}
\end{table*}

\begin{table*}
\begin{center}
 \setlength{\abovecaptionskip}{0.2cm}
\caption{The performance comparison on SemanticKITTI multiple scans benchmark. The item with arrow indicates the moving class} %
\label{table2}%
\setlength{\tabcolsep}{1.2pt}
\renewcommand{\arraystretch}{1.} 
% \addtolength{\tabcolsep}{-2.5pt}%
% \def\arraystretch{1.25}%   
\resizebox{\textwidth}{!}{
\begin{tabular}{ c | c | c| c | c c c c c c c c c c c c c c c c c c c c c c c c c}
  \hline
Methods & \rotatebox{90}{\textbf{mean-IoU}} & \rotatebox{90}{ {\textbf{FPS (Hz)}}}&\rotatebox{90}{ {\textbf{Params(M)}}} &  \rotatebox{90}{car } & \rotatebox{90}{bicycle } & \rotatebox{90}{motorcycle } & \rotatebox{90}{truck } & \rotatebox{90}{other-vehicle } & \rotatebox{90}{person } & \rotatebox{90}{bicyclist } & \rotatebox{90}{motorcyclist } & \rotatebox{90}{road } & \rotatebox{90}{parking } & \rotatebox{90}{sidewalk } & \rotatebox{90}{other-ground } & \rotatebox{90}{building } & \rotatebox{90}{fence } & \rotatebox{90}{vegetation } & \rotatebox{90}{trunk } & \rotatebox{90}{terrain } & \rotatebox{90}{pole } & \rotatebox{90}{traffic sign } & \rotatebox{90}{$\underrightarrow{\text{car}}$} & \rotatebox{90}{$\underrightarrow{\text{bicyclist}}$} & \rotatebox{90}{$\underrightarrow{\text{person}}$} & \rotatebox{90}{$\underrightarrow{\text{motorcyclist}}$} & \rotatebox{90}{$\underrightarrow{\text{other-vehicle}}$ } & \rotatebox{90}{$\underrightarrow{\text{truck}}$} \\
  \hline \hline
  TangentConv~\cite{tatarchenko2018tangent} & 34.1 &  {-}&0.4 & 84.9 & 2.0 & 18.2 & 21.1 & 18.5 & 1.6 & 0.0 & 0.0 & 83.9 & 38.3 & 64.0 & 15.3 & 85.8 & 49.1 & 79.5 & 43.2 & 56.7 & 36.4 & 31.2 & 40.3 & 1.1 & 6.4 & 1.9 & \textbf{30.1} & \textbf{42.2}\\
  DarkNet53Seg~\cite{behley2019semantickitti}  & 41.6 &  {-}&- & 84.1 & 30.4 & 32.9 & 20.2 & 20.7 & 7.5 & 0.0 & 0.0 & 91.6 & {64.9} & 75.3 &  {27.5}  & 85.2 & 56.5 & 78.4 & 50.7 & 64.8 & 38.1 & 53.3 & 61.5 & 14.1 & 15.2 & 0.2 & 28.9 & 37.8 \\
  SpSequenceNet~\cite{shi2020spsequencenet}  & 43.1 &  {3}&- & 88.5 & 24.0 & 26.2 & 29.2 & 22.7 & 6.3 & 0.0 & 0.0 & 90.1 & 57.6 & 73.9 & 27.1 & \textbf{91.2} & \textbf{66.8} & 84.0 & 66.0 & 65.7 & 50.8 & 48.7 & 53.2 & 41.2 & 26.2 & 36.2 & 2.3 & 0.1 \\
  TemporalLidarSeg~\cite{duerr2020lidar} & 47.0 &  {30}&- & {92.1} & 47.7 & 40.9 & \textbf{39.2} & {35.0} & 14.4 & 0.0 & 0.0 & {91.8} & 59.6 & {75.8} & 23.2 & 89.8 & 63.8 & 82.3 & 62.5 & 64.7 & 52.6 & 60.4 & 68.2 & 42.8 & 40.4 & 12.9 & 12.4 & 2.1 \\
  
   {TemporalLatticeNet~\cite{schutt2022abstract}}  &  {47.1} &  {6.5}&- &  {91.6} &  {35.4} &  {36.1} &  {26.9} &  {23.0} &  {9.4} &  {0.0} &  {0.0} &  {91.5} &  {59.3} &  {75.3} &  {27.5} &  {89.6} &  {65.3} &  {\textbf{84.6}}  &  {66.7} &  {\textbf{70.4}} &  {57.2} & {60.4} &  {59.7} &  {41.7} &  {9.4} &  {\textbf{48.8}} &  {5.9} &   {0.0} \\
   
  Meta-RangeSeg~\cite{wang2022meta} &  {49.7} &  {22}&6.8 &  {90.8} &  {50.0} &  {49.5} &  {29.5} &  {34.8} &  {16.6} &  {0.0} &  {0.0} &  {90.8} &  {62.9}&  {74.8} &  {26.5} &  {89.8} &  {62.1} &  {82.8} &  {65.7} &  {66.5} &  {56.2} &  {\textbf{64.5}} &  {69.0} &  {60.4} &  {57.9} &  {22.0} &  {16.6} &  {2.6} \\

  \hline
  LENet(Ours) &  {\textbf{53.0}} &  \textbf{25.3}&4.7 &  {\textbf{92.4}} &  {\textbf{57.0}} &  {\textbf{52.1}} &  {38.5} &  {\textbf{47.0}} &  {\textbf{19.0}} &  {0.0} &  {\textbf{4.3}} &  {\textbf{92.0}} &  {\textbf{68.6}}&  {\textbf{77.3}} &  {\textbf{29.9}} &  {90.2} &  {63.4} &  {83.4} &  {\textbf{68.1}} &  {67.6} &  {\textbf{58.0}} &  {62.9} &  {\textbf{75.2}} &  {\textbf{65.2}} &  {\textbf{62.6}} &  {22.5} &  {25.7} &  {2.0} \\
  \hline  
\end{tabular}}%
\end{center}
\vspace{-6mm}
\end{table*}%

In this work, we train the proposed neural network with three different loss functions, namely weighted cross-entropy loss ${\mathcal{L}}_{wce}$, Lov\'asz loss ${\mathcal{L}}_{ls}$ and boundary loss ${\mathcal{L}}_{bd}$. Finally, our total loss is following:
\begin{align}
\label{loss}
    {\mathcal{L}}=w_1{\mathcal{L}}_{wce}+w_2{\mathcal{L}}_{ls}+ w_3{\mathcal{L}}_{bd},
\end{align}
$w_1$, $w_2$ and $w_3$ are the weights with respect to each loss function. In our implementation, we set $w_1 = 1$, $w_2 = 1.5$ and $w_3 = 1$.

Three loss functions account for three different problems. To cope with the imbalanced classes problem, the weighted cross-entropy loss ${\mathcal{L}}_{wce}$ is employed to maximize the prediction accuracy for point labels, which is able to balance the distributions among different classes. It's defined as 
\begin{equation}
\mathcal{L}_{wce}(y, \hat{y})=-\sum_{i} \frac{1}{\sqrt{f_{i}}}  p\left(y_{i}\right) \log \left(p\left(\hat{y}_{i}\right)\right),
\end{equation}
where $y_i$ represents the ground truth, and $\hat{y}_i$ is prediction and $f_i$ is the frequency of the $i_{th}$ class. 

To solve the problem of optimizing the intersection-over-union(IoU), the Lov\'asz loss ${\mathcal{L}}_{ls}$ is used to maximize the intersection-over-union (IoU) score that is commonly used to in performance evaluation on semantic segmentation. It's defined as:
\begin{equation}
    \mathcal{L}_{ls}=\frac{1}{|C|} \sum_{c \in C} \overline{\Delta_{J_{c}}}(m(c)),\  m_{i}(c)\!=\!\left\{\begin{array}{ll}1\!-\!x_{i}(c) & \text{if } c\!=\!y_{i}(c) \\
x_{i}(c) & \text{otherwise}\end{array} \right.
\end{equation}
where $|C|$ is the class number, $\overline{\Delta_{J_{c}}}$ represents the Lov\'{a}sz extension of the Jaccard index, $x_{i}(c) \in[0,1]$ and $y_{i}(c) \in \{-1,1\}$ hold the predicted probability and ground truth label of pixel~$i$ for class $c$, respectively.

To account for the blurred segmentation boundaries problem as suggested in~\cite{razani2021lite, wang2022meta, cheng2022cenet}, the boundary loss function ${\mathcal{L}}_{bd}$ is used for LiDAR semantic segmentation, which can be formulated defined as follows:
\begin{equation}
\mathcal{L}_{bd}(y, \hat{y})=1-\frac{2 P_b^{c} R_b^{c}}{P_b^{c}+R_b^{c}},
\end{equation}
where $P_b^c$ and $R_b^c$ define the precision and recall of predicted boundary image $\hat{y}^{b}$ to real one $y^{b}$ for class $c$. The boundary image is computed as follows:
\begin{equation}
\begin{aligned}
&y^{b}=pool\left(1-y, \theta_{0}\right)-\left(1-y\right) \\
&\hat{y}^{b}=pool\left(1-\hat{y}, \theta_{0}\right)-\left(1-\hat{y}\right)
\end{aligned}
\end{equation}
where $pool(\cdot, \cdot)$ employs a pixel-wise max-pooling on a sliding window of size $\theta_{0}$. In addition, we set $\theta_{0}=3$.

\begin{figure*}
\setlength{\abovecaptionskip}{-0.2cm}
 \centering
\includegraphics[width=1.0 \linewidth,height=69mm]{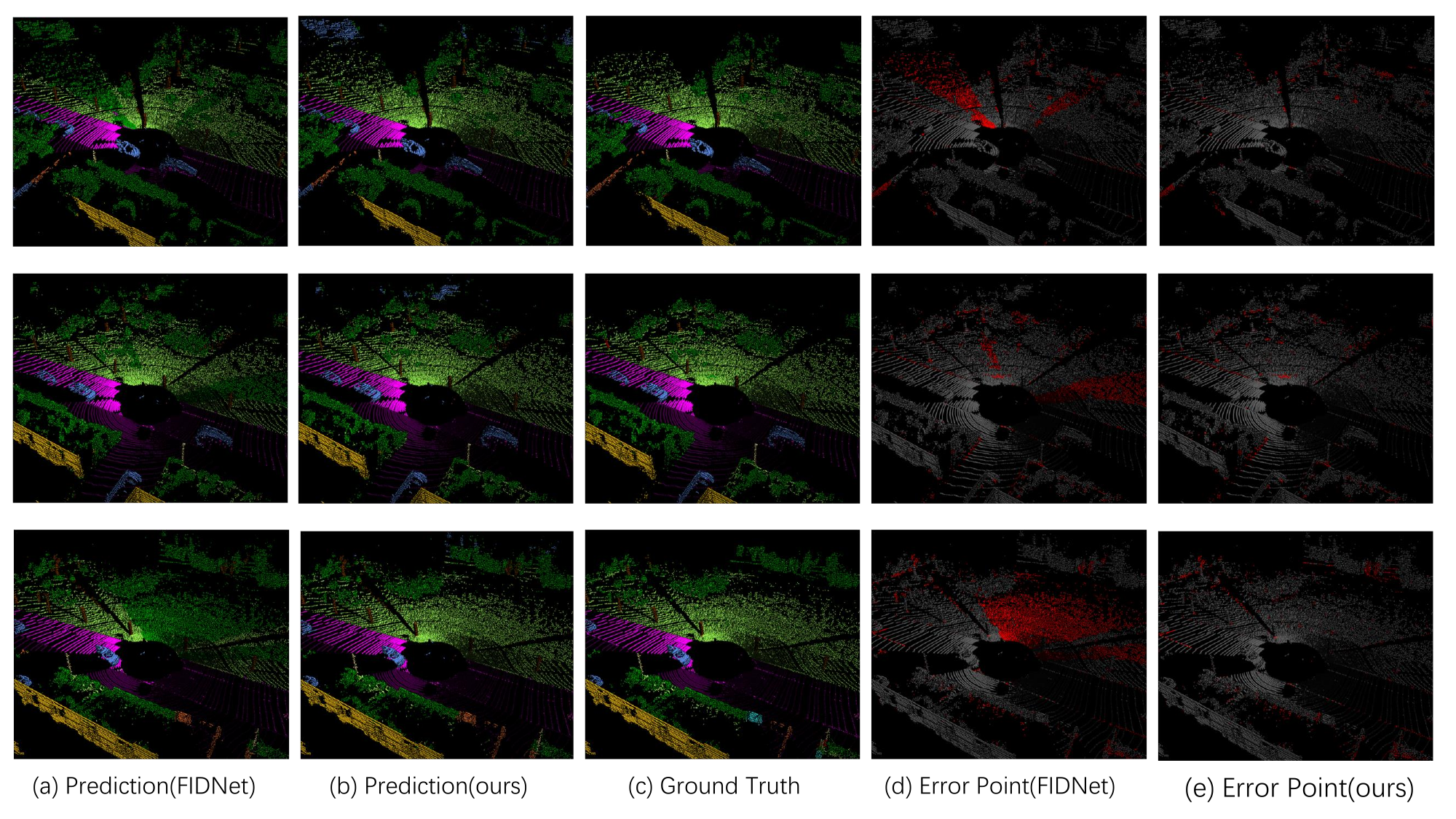}
    \caption{Qualitative analysis on the SemanticKITTI validation set (sequence 08). Where \textbf{(a)} and \textbf{b} are the predictions of FIDNet and our method respectively, \textbf{c} is the semantic segmentation ground truth, \textbf{d} and \textbf{e} are segmentation error maps 
 of FIDNet and our methods, where red indicates wrong prediction.}
    \label{fig:four}
    \vspace{-4mm}
\end{figure*}

In our proposed network, the classification head fuses the last three different dimensional feature map to conduct the output, which causes it's performance mainly relies on the last three IAC module. Therefore, we use the auxiliary segmentation head to further refine our proposed network accuracy. These auxiliary segmentation heads compute the weighted loss together with the main loss. Meanwhile, each loss has the corresponding weight since the different dimensional feature maps have different expressive power, which is different from ~\cite{xu2020squeezesegv3, cheng2022cenet}. The final loss function can be defined as,
\vspace{-1ex}
\begin{equation}
\label{eq: final_loss}
L_{total}=L_{main} + \sum_{i=1}^{3} \lambda_i L(y_{i},\hat{y}_{i})
\vspace{-1ex}
\end{equation}
where $L_{main}$ is the main loss, $y_{i}$ is the semantic output obtained from stage $i$, and $\hat{y}_{i}$ represents the corresponding semantic label. $L(\cdot)$ is computed according to Equation~\ref{loss}. In our implementation, we set $\lambda_1 = 1$, $\lambda_2 = 1$ and $\lambda_3 = 0.5$, empirically.

\section{EXPERIMENT}

\begin{table}[t]\tiny
\vspace{-2mm}
\centering
\setlength{\abovecaptionskip}{0.1cm}
\caption{Evaluation results on the SemanticPOSS test split.}
\label{tab:poss}
\setlength{\tabcolsep}{2.pt}
\renewcommand{\arraystretch}{1.} 
	\resizebox{\linewidth}{!}{
		\begin{tabular}{c|ccccccccccccc|c}
			\hline
			& 
			\rotatebox{90}{person} & 
			\rotatebox{90}{rider} & 
			\rotatebox{90}{car} & 
			\rotatebox{90}{truck} & 
			\rotatebox{90}{plants} & 
			\rotatebox{90}{traffic sign} & 
			\rotatebox{90}{pole} & 
			\rotatebox{90}{trashcan} & 
			\rotatebox{90}{building} & 
			\rotatebox{90}{cone/stone} & 
			\rotatebox{90}{fence} & 
			\rotatebox{90}{bike} &
			\rotatebox{90}{ground} & 
			\rotatebox{90}{mIoU}
			\\
			\hline
			SqueezeSeg \cite{wu2018squeezeseg} & 14.2 & 1.0 & 13.2 & 10.4 & 28.0 & 5.1 & 5.7 & 2.3 & 43.6 & 0.2 & 15.6 & 31.0 & 75.0 & 18.9\\
			SqueezeSeg + CRF \cite{wu2018squeezeseg} & 6.8 & 0.6 & 6.7 & 4.0 & 2.5 & 9.1 & 1.3 & 0.4 & 37.1 & 0.2 & 8.4 & 18.5 & 72.1 & 12.9\\
			SqueezeSegV2 \cite{wu2019squeezesegv2} & 48.0 & 9.4 & 48.5 & 11.3 & 50.1 & 6.7 & 6.2 & 14.8 & 60.4 & 5.2 & 22.1 & 36.1 & 71.3 & 30.0 \\
			SqueezeSegV2 + CRF \cite{wu2019squeezesegv2} & 43.9 & 7.1 & 47.9 & 18.4 & 40.9 & 4.8 & 2.8 & 7.4 & 57.5 & 0.6 & 12.0 & 35.3 & 71.3 & 26.9\\
			RangeNet53 \cite{milioto2019rangenet++} & 55.7 & 4.5 & 34.4 & 13.7 & 57.5 & 3.7 & 6.6 & 23.3 & 64.9 & 6.1 & 22.2 & 28.3 & 72.9 & 30.3\\
			RangeNet53 + KNN \cite{milioto2019rangenet++} & 57.3 & 4.6 & 35.0 & 14.1 & 58.3 & 3.9 & 6.9 & 24.1 & 66.1 & 6.6 & 23.4 & 28.6 & 73.5 & 30.9\\
			MINet \cite{li2020multi} & 61.8 & 12.0 & 63.3 & 22.2 & 68.1 & 16.3 & 29.3 & 28.5 & 74.6 & 25.9 & 31.7 & 44.5 & 76.4 & 42.7\\
			MINet + KNN \cite{li2020multi} & 62.4 & 12.1 & 63.8 & 22.3 & 68.6 & 16.7 & 30.1 & 28.9 & 75.1 & 28.6 & 32.2 & 44.9 & 76.3 & 43.2 \\
			FIDNet \cite{zhao2021fidnet} & 75.6 & 20.3 & 76.4 & 24.3 & 72.4 & 18.3 & 31.0 & 39.3 & 78.2 & 39.2 & 43.8 & 52.5 & 80.5 & 50.1 \\
			FIDNet + KNN \cite{zhao2021fidnet} & 76.0 & 20.9 & 77.2 & 24.5 & 72.6 & 18.8 & 31.9 & 41.4 & 78.6 & \textbf{41.1} & 43.4 & 51.8 & 79.5 & 50.6 \\
            CENet \cite{cheng2022cenet} & 75.4 & 24.6 & 80.0 & 27.5 & 71.6 & 25.4 & 32.4 & 51.8 & 77.3 & 34.5 & \textbf{49.7} & 52.7 & \textbf{80.8} & 52.6 \\
			CENet + KNN \cite{cheng2022cenet} & 76.0 & 24.7 & 80.7 & 27.6 & 71.6 & 25.5 & 33.1 & 53.5 & 77.7 & {37.5} & 48.8 & 51.9 & 79.7 & 53.0 \\
			%		FIDNet-Point \cite{zhao2021fidnet} & 74.8 & 20.0 & 75.7 & 21.7 & 71.5 & 28.4 & 29.9 & 30.0 & 77.3 & 38.9 & 46.5 & 53.2 & 80.2 & 49.9 \\
			%		FIDNet-Point + KNN \cite{zhao2021fidnet} & 75.0 & 20.1 & 76.3 & 21.8 & 71.7 & 28.8 & 30.4 & 30.9 & 77.7 & 40.6 & 46.4 & 52.8 & 79.7 & 50.2 \\
			\hline
			Ours & 76.8 & 26.6 & 81.1 & \textbf{31.8} & 73.9 & 26.4 & 34.4 & 53.8 & 80.4 & 29.4 & 50.5 & \textbf{52.8} & 79.2 & 53.6 \\
			Ours + KNN & \textbf{77.0} & \textbf{26.8} & \textbf{81.5} & \textbf{31.8} & \textbf{74.0} & \textbf{26.5} & \textbf{34.9} & \textbf{55.0} & \textbf{80.7} & 31.6 & \textbf{49.7} & 52.1 & 78.2 & \textbf{53.8} \\
			%		Ours & 76.0 & 24.2 & 79.2 & 26.7 & 72.3 & 19.6 & 30.9 & 48.3 & 77.3 & 37.4 & 50.1 & 52.5 & 80.7 & 51.9 \\
			%		Ours + KNN & 76.4 & 24.3 & 79.6 & 26.8 & 72.5 & 19.7 & 31.3 & 49.2 & 77.5 & 40.3 & 50.0 & 52.2 & 80.2 & 52.3 \\
			\hline
	\end{tabular}}
	\vspace{-5mm}
\end{table}

\begin{table}[t]\tiny
\vspace{2mm}
\centering
\setlength{\abovecaptionskip}{0.1cm}
\caption{Evaluation results on the nuScenes validation dataset.}
\label{tab:nuscene}
\setlength{\tabcolsep}{2.pt}
\renewcommand{\arraystretch}{1.} 
	\resizebox{\linewidth}{!}{
		\begin{tabular}{c|cccccccccccccccc|c}
			\hline
			& 
			\rotatebox{90}{barrier} & 
			\rotatebox{90}{bicycle} & 
			\rotatebox{90}{bus} & 
			\rotatebox{90}{car} & 
			\rotatebox{90}{construction vehicle} & 
			\rotatebox{90}{motorcycle} & 
			\rotatebox{90}{pedestrian} & 
			\rotatebox{90}{traffic cone} & 
			\rotatebox{90}{trailer} & 
			\rotatebox{90}{truck} & 
			\rotatebox{90}{driveable surface} & 
			\rotatebox{90}{other flat} &
			\rotatebox{90}{sidewalk} & 
			\rotatebox{90}{terrian} &
                \rotatebox{90}{manmade} &
			\rotatebox{90}{vegetation} & 
			\rotatebox{90}{mIoU}
			\\
			\hline
			FIDNet \cite{zhao2021fidnet} & 73.2 & 38.2 & 86.9 & 83.0 & 38.9 & 71.2 & 70.9 & 60.0 & 66.7 & 76.3 & 93.3 & 70.3 & 72.9 & 72.4 & 86.6 & 84.9 & 71.7 \\
			FIDNet + KNN \cite{zhao2021fidnet} & 57.1 & 35.6 & 87.3 & 82.2 &40.7 & 71.2 & 72.5 & 62.3 & 68.0 & 76.3 & 96.1 & 71.4 & 73.6 & 73.4 & 87.8 & 86.1 & 71.4 \\
                CENet \cite{cheng2022cenet} & 74.8 & 39.9 & 85.6 & 83.6 & 46.3 & \textbf{78.5} & 72.1 & 60.8 & 66.4 & 77.3 & 93.1 & 69.5 & 73.0 & 72.2 & 86.4 & 84.4 & 72.7 \\
			CENet + KNN \cite{cheng2022cenet} & 58.4 & 37.4 & 86.1 & 82.8 &46.9 & 78.0 & 73.7 & 62.7 & 67.3 & 77.3 & 95.9 & 70.7 & 73.6 & 73.2 & 87.5 & 85.6 & 72.3 \\
			\hline
			Ours & \textbf{76.0} & \textbf{41.1} & 88.1 & \textbf{84.6} & 48.8 & 78.0 & 73.2 & 62.3 & 67.8 & \textbf{79.4} & 93.4 & 72.1 & 73.7 & 72.7 & 87.1 & 85.3 & \textbf{74.0} \\
			Ours + KNN & 59.2 & 28.3 & \textbf{88.6} & 83.8 & \textbf{49.7} & 77.3 & \textbf{74.6} & \textbf{64.3} & \textbf{68.8} & \textbf{79.4 }& \textbf{96.3 }& \textbf{73.3} & \textbf{74.3 }& \textbf{73.7 }& \textbf{88.1} & \textbf{86.3} & 73.5  \\
			\hline
	\end{tabular}}
	\vspace{-5mm}
\end{table}

\subsection{Experiment Setups}
To evaluate our method, we use SemanticKITTI, which is a large-scale dataset for the task of point cloud segmentation of autonomous driving scenes. It provides the dense point-wise annotation for 22 sequences (43,551 scans) in KITTI Odometry Benchmark. Sequence 00 to 10 (19,130 scans) are used for training, 11 to 21 (20,351 scans) for testing. We follow the setting in \cite{behley2019semantickitti}, and use sequence 08 (4,071 scans) for validation. To evaluate the effectiveness of our proposed approach, we submit the output the online evaluation website to obtain the results on the testing set. 

To verify the universality of our proposed method, we conducted additional experiments on SemanticPOSS and nuScenes datasets. SemanticPOSS, which was collected at Peking University, uses the same data format as SemanticKITTI. However, compared to SemanticKITTI, it is a smaller and sparser dataset with 2,988 complex LiDAR scans and a large number of dynamic instances that make it a challenging benchmark. On the other hand, nuScenes consists of 1000 sequences, each with 20 seconds of duration, recorded with a 32-beam LiDAR sensor. Among the three datasets, nuScenes has the fewest points per frame in its point cloud, but the greatest number of frames.

To faciliate the fair comparison, we evaluate the performance of different methods with respect to the mean intersection over union metric (mIoU), which is defined as below:
\begin{align}
	\text{mIoU} = \frac{1}{n}\sum_{c=1}^{n}\frac{\text{TP}_c}{\text{TP}_c + \text{FP}_c + \text{FN}_c}, \label{eq:miou}
\end{align}
where $\text{TP}_c$, $\text{FP}_c$, and $\text{FN}_c$ represent true positive, false positive, and false negative predictions for the  class $c$.

\subsection{Implementation details.}
We implemented our proposed method using PyTorch\cite{paszke2017automatic} and conducted all our experiments on a computer with four NVIDIA RTX 3090 GPUs. To train the network, we used a batch size of 8 with a range image size of $64 \times 2048$. We employed the AdamW\cite{loshchilov2017decoupled} optimizer with default settings in PyTorch, and the initial learning rate was set to 2$e^{-3}$, which we dynamically adjusted using a cosine annealing scheduler over 50 epochs. During the training process, we performed data augmentation using random rotation, random point dropout, and flipping the 3D point cloud.

\subsection{Evaluation Results and Comparisons}
Table~\ref{table1} displays the results of various recent methods tested on the SemanticKITTI single scan benchmark. Our presented method achieves the state-of-the-art performance compared to both point-based and image-based methods on the single scan benchmark (64.5\% mIoU). Table~\ref{table2} displays the quantitative results of various recent methods on the SemanticKITTI multiple scans benchmark. Our presented methods also achieve state-of-the-art performance with 53.0\% mIoU. It is worth noting that our proposed network is lightweight, with just around 4.7 million parameters and operates fast at approximately 26 frames per second while maintaining high accuracy.

Table~\ref{tab:poss} displays a comparison of our proposed LENet with other related works on SemanticPOSS. Due to differences in sensors and environments, all methods achieve slightly lower accuracy. Despite this, our proposed method achieves state-of-the-art performance, as demonstrated by its high scores in both overall mIoU and nearly all class-wise mIoU metrics.

Table~\ref{tab:nuscene} compares the performance of our proposed LENet against FIDNet and CENet on the nuScenes validation set. Compared to SemanticKITTI and SemanticPOSS, the nuScenes dataset has the most sparsely sampled point cloud, making the task more challenging. Despite this, our method still achieves high accuracy, surpassing both the baseline FIDNet and CENet by 2.3\% and 0.4\%, respectively.

In order to better visualize the improvements of our proposed model over the baseline system, we present a qualitative comparison of FIDNet and LENet in Fig.~\ref{fig:four}. We compare the prediction results and generated error maps of both models on three data frames from the SemanticKITTI validation set. The comparison demonstrates a significant improvement of our proposed method over FIDNet.

\subsection{Ablation Studies}

\begin{table}[t]
	\vspace{-2mm}
	\centering
        \setlength{\abovecaptionskip}{0.1cm}
	\caption{Ablation study evaluated on SemanticKITTI validation set.}
	\label{tab:ablation}
	\resizebox{\linewidth}{!}{
		\begin{tabular}{|c|c|cccc|c|c|} 
			\hline
                \multicolumn{1}{|c|}{\textbf{Baseline}} & 
                \multicolumn{1}{|c|}{\textbf{Row}} & 
                \multicolumn{1}{c}{\textbf{MSCA}} & 
                \multicolumn{1}{c}{\textbf{IAC}} &
                \multicolumn{1}{c}{\textbf{\begin{tabular}[c]{@{}c@{}}Boundary\\ Loss\end{tabular}}} &
                \multicolumn{1}{c|}{\textbf{\begin{tabular}[c]{@{}c@{}}Auxiliary\\ Loss\end{tabular}}} &
                \multicolumn{1}{c|}{\textbf{mIou}} &
                \multicolumn{1}{c|}{\textbf{Params(M)}} \\ 
			\hline
			FIDNet &1 &\xmark &\xmark  & \xmark &\xmark &  58.3 & 6.05 \rule{0pt}{3ex} \\ 
			\hline
			\multirow{5}{*}{Ours} &2&\cmark &\xmark &\xmark  &\xmark  &59.3 & 5.28 \rule{0pt}{3ex} \\ 
			\cline{2-8}
			&3& \cmark &\cmark &\xmark &\xmark & 59.9 & 4.74 \rule{0pt}{3ex} \\ 
			\cline{2-8}
			&4&\cmark & \cmark &\cmark &\xmark  &  61.1 & 4.74 \rule{0pt}{3ex} \\ 
			\cline{2-8}
			&5&\cmark &\cmark & \cmark &\cmark  &  63.1 & 4.74 \rule{0pt}{3ex} \\ 
			\hline
	\end{tabular}}
	\vspace{-5mm}
\end{table}

In this section, we present the results of multiple ablation experiments conducted on the SemanticKITTI validation set (sequence 08) to observe the performance impact of individual modules in our proposed network. For a fair comparison, we use FIDNet as the baseline since it has a similar network structure to ours. We then replace the baseline with our proposed MSCA and IAC modules from Section~\ref{method}. Finally, we incrementally add the boundary loss and auxiliary loss to evaluate their effectiveness for our network. Table~\ref{tab:ablation} displays the number of model parameters and corresponding mIoU scores on the SemanticKITTI validation set. Our proposed LENet approach demonstrates over 4.8\% improvement and 25\% reduction in parameters compared to the baseline, validating the efficacy of each individual module.

Our proposed network achieved improved accuracy, with an increase of over $1.0\%$ when we replaced the basic block of the baseline with the MSCA. Meanwhile, because of the use of group convolution in MSCA, our network reduced the total number of parameters to 5.28M, making it lightweight. After replacing the original decoder with the IAC, our network also yielded an improved accuracy, which was approximately $0.6\%$ better compared to the prev. Despite IAC containing learnable convolutional parameters, the improved performance clearly demonstrates the efficacy of ICA. Furthermore, the use of ICA eliminates the need for a heavy segmentation head in the decoder, like FIDNet, resulting in a smaller number of parameters (4.74M). In terms of the additional loss functions, the boundary loss delivered a performance gain of over $1.2\%$, while the auxiliary loss achieved an impressive performance jump of over $2.0\%$.

\section{CONCLUSIONS}
We propose a lightweight and efficient real-time CNN model called LENet for the task of LiDAR point cloud segmentation. Our novel encoder-decoder architecture is based on MSCA and IAC. In addition, we incorporate a boundary loss to emphasize semantic boundaries and several extra segmentation heads to further improve feature learning ability without affecting parameter or efficiency costs. Our proposed method achieves state-of-the-art performance on single-scan semantic segmentation and multiple-scan semantic segmentation as well as real-time capability, as demonstrated by the evaluation on the SemanticKITTI test dataset. Moreover, we conducted further experiments on SemanticPOSS and nuScenes and achieved promising results.


%%%%%%%%%%%%%%%%%%%%%%%%%%%%%%%%%%%%%%%%%%%%%%%%%%%%%%%%%%%%%%%%%%%%%%%%%%%%%%%%

%%%%%%%%%%%%%%%%%%%%%%%%%%%%%%%%%%%%%%%%%%%%%%%%%%%%%%%%%%%%%%%%%%%%%%%%%%%%%%%%

%%%%%%%%%%%%%%%%%%%%%%%%%%%%%%%%%%%%%%%%%%%%%%%%%%%%%%%%%%%%%%%%%%%%%%%%%%%%%%%%

%%%%%%%%%%%%%%%%%%%%%%%%%%%%%%%%%%%%%%%%%%%%%%%%%%%%%%%%%%%%%%%%%%%%%%%%%%%%%%%%

% References are important to the reader; therefore, each citation must be complete and correct. If at all possible, references should be commonly available publications.

% \vfill\pagebreak

\bibliographystyle{IEEEtran}
\bibliography{IEEEabrv,IEEEexample}

\end{document}